\documentclass[conference]{IEEEtran}
\IEEEoverridecommandlockouts

\usepackage{cite}
\usepackage{amsmath,amssymb,amsfonts}
\usepackage{algorithmic}
\usepackage{graphicx}
\usepackage{textcomp}
\usepackage{xcolor}
\def\BibTeX{{\rm B\kern-.05em{\sc i\kern-.025em b}\kern-.08em
    T\kern-.1667em\lower.7ex\hbox{E}\kern-.125emX}}
\begin{document}

\title{Detecting radar targets swarms in range profiles with a partially complex-valued neural network}

\author{\IEEEauthorblockN{Martin Bauw}
\IEEEauthorblockA{\textit{DEMR, ONERA} \\
\textit{Université Paris-Saclay}\\
91120 Palaiseau, France \\
martin.bauw@onera.fr}
}

\maketitle

\begin{abstract}
Correctly detecting radar targets is usually challenged by clutter and waveform distortion. An additional difficulty stems from the relative proximity of several targets, the latter being perceived as a single target in the worst case, or influencing each other's detection thresholds. The negative impact of targets proximity notably depends on the range resolution defined by the radar parameters and the adaptive threshold adopted. This paper addresses the matter of targets detection in radar range profiles containing multiple targets with varying proximity and distorted echoes. Inspired by recent contributions in the radar and signal processing literature, this work proposes partially complex-valued neural networks as an adaptive range profile processing. Simulated datasets are generated and experiments are conducted to compare a common pulse compression approach with a simple neural network partially defined by complex-valued parameters. Whereas the pulse compression processes one pulse length at a time, the neural network put forward is a generative architecture going through the entire received signal in one go to generate a complete detection profile.
\end{abstract}


\section{Introduction}

Radar range profiles (RP) are commonly processed using matched filters and constant false alarm rate (CFAR) tests to detect the presence of at least one target within range bins. The matched filter (MF) is the optimal filter to find a template signal, which in active radars corresponds to the transmitted waveform, under the hypothesis of additive white Gaussian noise. A common CFAR test relies on the averaging of neighboring range cells, whose value is computed using a matched filter, to define an adaptive waveform echo detection threshold.  Such a test is called the cell-averaging CFAR (CA-CFAR), and is one of the numerous existing CFAR tests, different approaches being required to tackle various kinds of clutter environments and targets densities \cite{rohling1983radar} \cite{gandhi1988analysis}. As for the CA-CFAR test, the MF is not ideal in typical use cases, for instance when a weak target is masked by a strong one in a sidelobe. The MF can thus be complemented by mismatched filters (MMF), which may be more relevant depending on the radar task considered. The use of both MF and MMF is a reminder of the impossibility of an optimum radar waveform for target resolution \cite{rihaczek1965radar}. The challenges of neighboring targets, sidelobes echoes and limited resolutions are of growing importance as radar targets will increasingly appear in groups of varying radar cross section (RCS). Drones can fly in swarms or waves in search for aesthetic or air defense saturation effects, and next generation fighter aircrafts are expected to team up with loyal wingman unmanned aerial vehicles (UAV). Such challenges motivate this work, though the latter does not address typical swarm-specific issues such as configuring radars to yield the range and angle resolutions necessary to separate individual targets within a swarm, or defining radar hits clustering strategies. 

The need to adjust baseline methods such as the MF and the CA-CFAR test led to the use of least-square estimation, constrained optimization and iterative gradient descent to build adaptive alternatives \cite{blunt2006adaptive} \cite{rabaste2015mismatched} \cite{mccormick2016gradient}. The distortion of radar echoes may, in some cases, go beyond thermal noise, clutter and Doppler shift. For instance, targets coated with plasma may additionally and deliberately disturb the reflected waveform for improved stealth \cite{xu2017evaluations}. The numerous and diverse distortion causes suggest particularly flexible adaptive radar echoes processing should be evaluated, opening the way to machine learning-based radar processing, which comes with the drawback of requiring a substantial amount of training and validation data. Relying on machine learning, including neural networks, has become a relevant option for various radar systems functions including radar targets detection \cite{akhtar2018neural} \cite{diskin2024cfarnet}. Automatic threat detection is crucial as a delayed reaction can easily lead to the system's destruction, and the availability of completely or partially complex-valued neural networks and backpropagation conveniently adapt neural networks to radar IQ data \cite{smith2024frequency} \cite{lei2024understanding} \cite{brooks2019hermitian}. 


This paper puts forward a radar range profile processing where the normalized pulse compression followed by a CA-CFAR detection and its sequential application over segments of the received signal is replaced by a single neural network. Similar works thus include \cite{akhtar2018neural} which relies on traditional pulse compression to preprocess echoes then fed to a neural network detector, the use of a neural network to preprocess a range-Doppler map before CA-CFAR detection in \cite{pervdoch2024improved}, and \cite{gandhi1997neural} which detects a real-valued signal in non-Gaussian noise using a real-valued neural network. The originality of this work stems from the combination of inherently complex-valued neural network parameters with varying waveform bandwidth distorting received echoes, targets swarms to densely populate range profiles and the use of raw IQ signal instead of the matched filter output as neural network input. The paper first introduces the artificial data generation method in section \ref{datasets-generation}. Section \ref{range-profile-processing} then details the neural network-based range profile processing along with the pulse compression baseline, and experimental results are finally presented in section \ref{experiments}.

\section{Datasets generation}
\label{datasets-generation}

A large quantity of radar range profiles is generated using different signal parameters and varying targets positions. The training, validation and test sets include both single-target and multiple-targets range profiles. The training set range profiles exclusively contain regularly spaced targets whereas the test set range profiles exclusively contain irregularly spaced targets. The "baseline" training set is generated using a single bandwidth $B_{train}$, used to define the matched filter, and relatively low-level noise. The "enriched" training set additionally carries reduced bandwidth echoes, along with empty and contrastive range profiles. Empty range profiles are noise-only range profiles, while contrastive range profiles carry unmodulated sine pulses echoes. Such enrichments aim at providing the neural network with more diverse target-free range bins and contrastive information to learn from. The test set is generated with a different set of bandwidths, including the training set "true" bandwidth $B_{train}$ defining the matched filter, different noise levels, different pulse reflection coefficients and irregularly spaced targets. The irregularly spaced targets stem from regularly spaced training set targets positions randomly shifted. This way three types of waveform distortions are taken into account to create relevant radar pulse echoes : waveform bandwidth modification, received waveform amplitude and increased noise levels. The bandwidth modification will be a bandwidth reduction, such a distortion thus differing from a mere Doppler shift, as for the latter the bandwidth covered remains the same. The waveform selected to generate target echoes is the linear frequency modulated (LFM) chirp, which is defined by the following equation and is a common choice for radar applications \cite{akhtar2023high} \cite{xu2017evaluations} \cite{rabaste2015mismatched}:

\begin{equation}
\label{lfm-chirp}
s(t) = \exp\left(j\pi\dfrac{B_{train}}{T}t^2\right)
\end{equation}

where $T$ stands for the chirp duration. Let us now assume this waveform is defined by $n$ samples, the transmitted waveform is thus actually:

\begin{equation}
\label{lfm-chirp-sampled}
\boldsymbol{s} = \left(s_1 \hdots s_n\right)
\end{equation}

The previous waveform is repeated, with the previously listed distortions, along the range axis to simulate the signal received over a single pulse repetition interval (PRI) following the initial pulse transmission. Thousands of training and test range profiles can be generated using the three types of waveform distortions mentioned thanks to programming loops creating range profiles for the numerous targets shifts and targets numbers allowed by the range profile size. Target echoes are spaced using a limited stride in order to produce intersecting matched filters and interfering contributions to CA-CFAR thresholds. While diversely distributed isolated single-target range profiles do not offer valuable diversity for MF processing happening at pulse-scale, they do for the neural network processing as the latter considers an entire range profile at a time, making absolute targets locations relevant.

\section{Range profile processing}
\label{range-profile-processing}

\begin{figure}[!t]
\centering
\includegraphics[width=3.5in]{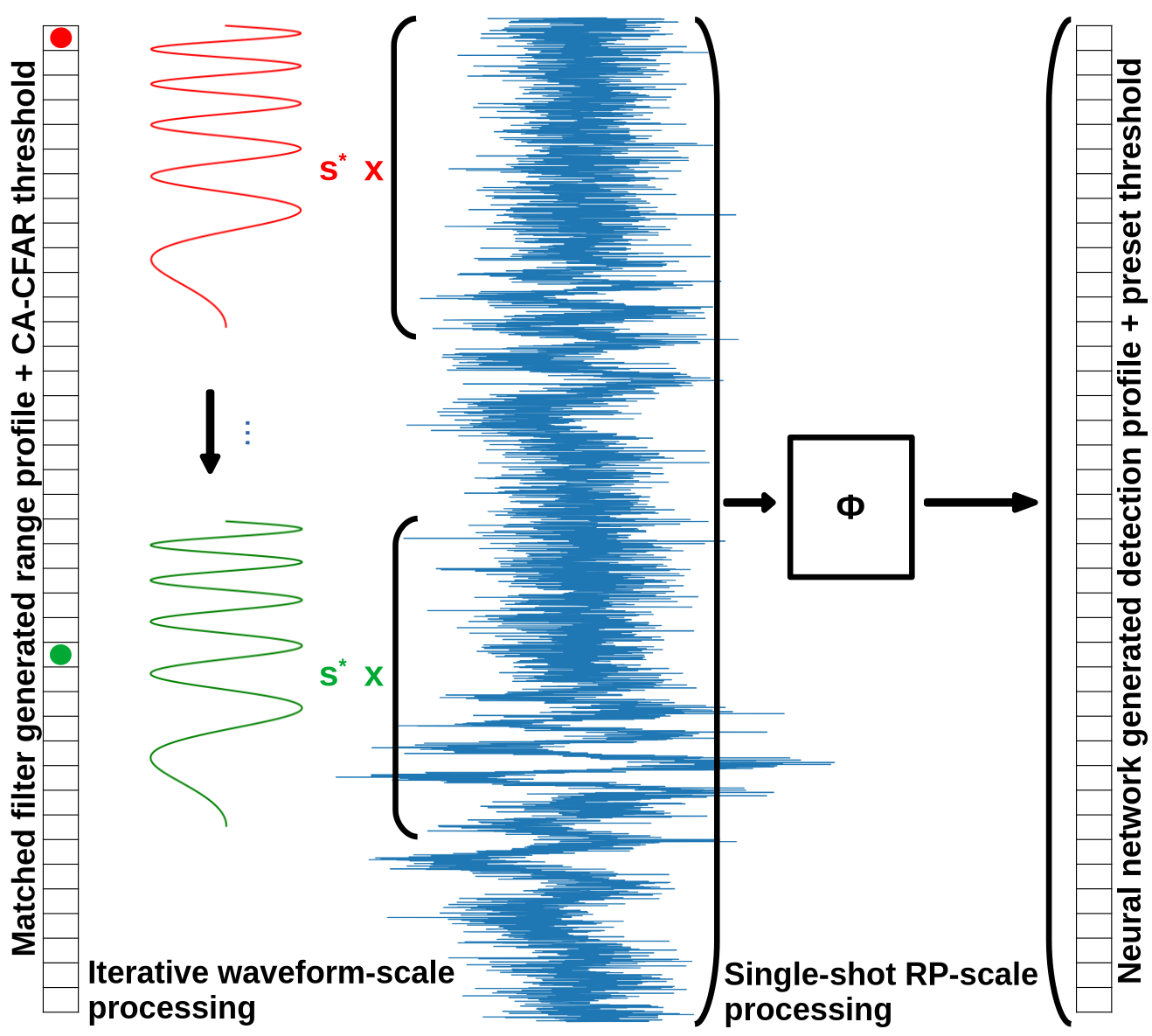}
\caption{Matched filter range profile processing and RX-scale range profile processing: the MF implies applying the same filter sequentially over shifted slices of the range profile, whereas the proposed RX-scale range profile processing considers the entire range profile all at once using the neural network $\Phi$.}
\label{RX_scale_processing}
\end{figure}

This paper proposes replacing both the matched filter and the CA-CFAR threshold with a neural network. Therefore, the baseline we compare our proposed method to is the matched filter itself with a downstream CA-CFAR detection characterized by a $P_{fa}$ which allows for the computation of the threshold factor $\alpha$ multiplying the $R$ reference cells average to produce the local detection threshold applied to detection test range bin values:

\begin{equation}
\label{cfar-multiplier}
\alpha = R \left(P_{fa}^{-\frac{1}{R}}-1 \right)
\end{equation}

In the baseline case, the $i$-th cell in the range profile is thus defined, following the matched filter normalized by the transmitted pulse energy, by:


\begin{equation}
\label{rp-cell-mf-norm}
RP_{i} = \dfrac{1}{\sum_{k=1}^n \lvert s_k \rvert^2} \lvert\sum_{k=1}^n x_{i+k} s_k^{*}\rvert^2
\end{equation}

where $x_j$ is a received signal coefficient sample. The echo measures can then be converted into decibels, as it was done to produce the signal seen on Fig.~\ref{example_test_rp}:

\begin{equation}
\label{rp-cell-mf-norm-db}
\left(RP_{i}\right)_{dB} = 20log_{10}\left(\dfrac{1}{\sum_{k=1}^n \lvert s_k \rvert^2} \lvert\sum_{k=1}^n x_{i+k-1} s_k^{*} \rvert^2 \right)
\end{equation}

Successive $\left(RP_{i}\right)$ are stacked to form the output range profile $\boldsymbol{RP}$ spanning $m$ range bins and fed to the CA-CFAR detector $\Gamma$:

\begin{equation}
\label{rp-cell-mf-norm-db-ca-cfar}
\boldsymbol{\Gamma_{out}} = \Gamma \left( \left(\left(RP_{1}\right) \hdots \left(RP_{i}\right) \hdots \left(RP_{m}\right)\right)\right)
\end{equation}

It is important to note that the baseline approach based on the matched filter processes one pulse length of received signal at a time to create individual range profile cells values $\left(RP_{i}\right)$. A loop may thus be used to sweep over the received signal with the matched filter to create a complete range profile, although in reality this would rather be achieved in the Fourier space using the discrete Fourier transform. The $\left(RP_{i}\right)$ cells are locally gathered as guard and reference cells to conduct the local average comparison with the cell under test, eventually producing a complete $\boldsymbol{\Gamma_{out}}$. This differs from the neural network approach put forward, which takes the entire received signal $\boldsymbol{x}$ as input to produce a complete detection profile in one go. This fundamental difference is illustrated on Fig.~\ref{RX_scale_processing}. The proposed approach thus replaces both the detection of \eqref{rp-cell-mf-norm-db-ca-cfar} and the matched filter of \eqref{rp-cell-mf-norm} with the neural network $\Phi$. The latter directly translates the received signal into range bins  with values in $[0,1]$ and which are directly compared to an arbitrary threshold set to $0.5$.

\begin{figure}[!t]
\centering
\includegraphics[width=2.4in]{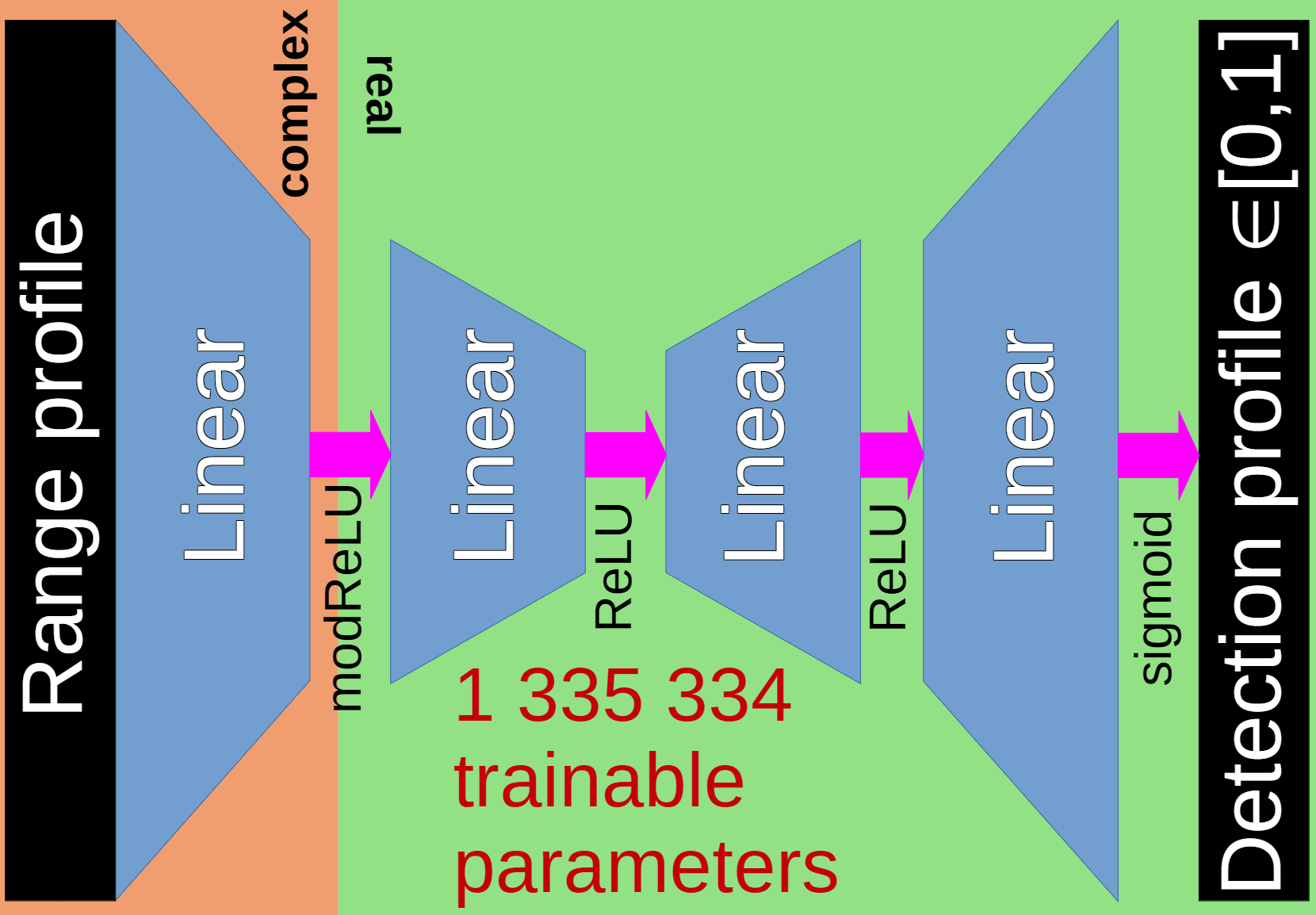}
\caption{RX-scale range profile processing using a hybrid real-complex-valued autoencoder. The proposed RX-scale range profile processing considers the entire range profile all at once using the neural network $\Phi$, and produces a detection profile immediately comparable to an arbitrary threshold.}
\label{NNet_arch_AE}
\end{figure}

The neural network $\Phi$ is partially complex-valued : the first linear layer along with the first activation function are complex-valued and trained using adapted backpropagation. Only the real part of the output of the first activation function is given to the following, real-valued, linear layer. The hybrid architecture thus defined is illustrated on Fig.~\ref{NNet_arch_AE}. Entirely complex-valued alternatives were considered but led to lower performances. The final operation implemented within the neural network is a real-valued sigmoid, yielding detection profile values restricted to $[0,1]$. The complex-valued activation function selected is the modReLU initially proposed in \cite{arjovsky2016unitary} and put forward in \cite{trabelsi2017deep}. It is defined as follows for any complex-valued input $z$ with angle $\theta$:

\begin{equation}
\label{moderelu}
modReLU(z) = ReLU(|z|+bias)e^{i\theta}
\end{equation}

The neural network output is a detection profile whose size is similar to the input range profile size, giving a generative function to the network, which amounts to an autoencoder. The range profile-scale of the neural network processing could be seen as an unfair advantage, as the neural network benefits from the entire range context to identify distorted echoes. In a way, this can be seen as pushing the number of cells available to a CFAR test to the maximum along the range axis. On the other hand, the neural network trained does not have direct access to the transmitted waveform used to define the matched filter weights. The latter is only accessible to the network through the distorted echoes carried by training set range profiles.

\section{Experiments}
\label{experiments}

The number of range profiles generated and used for the reported experiments along with the parameters of the datasets generation process are specified in Table \ref{data_gen_params}. The sampling frequency was set to $F_S = 2$ MHz, the number of range bins to $1000$ and the pulse length to $200$ bins, i.e. in \eqref{lfm-chirp-sampled} $n=200$. The CA-CFAR detection is defined by $G=4$ guard cells ($2$ on each side) and $R=20$ reference cells ($10$ on each side). The threshold factor defined in \eqref{cfar-multiplier} is configured to produce a false alarm probability $P_{fa}=10^{-3}$. The high targets density, an inherent feature of targets swarms processing, implies that neighboring targets contribute to heightened CA-CFAR thresholds.

One should note that the sampling frequency $F_S$ is theoretically required to be a multiple of the transmitted pulse bandwidth for CA-CFAR reference cells to possibly coincide with matched filter zeros so that only clutter and thermal noise contributes to the average defining the local clutter estimate. In our experiments this common assumption will not be verified since many of the test set range profiles carry echoes of improper, distorted, bandwidth. The CA-CFAR detection will nevertheless be conducted under the assumption of the transmitted bandwidth being equal to $B_{train}$, since in our case differing bandwidths amount to unpredictable distortions. The transmitted bandwidth assumption then allows for CA-CFAR to happen as it translates into the range bins shift required to pick the previously mentioned MF zeros. As indicated in Table \ref{data_gen_params} $F_S=2B_{train}$, which implies that for a given target one in two neighboring range bins constitutes a MF zero. Only these bins may be counted as guard or reference cells. One should additionally note that reported detection performances do not include all range bins: only range bins containing an actual target will be counted as positives, and empty range bins are ignored in detection counts used to compute $P_d$ and $P_{fa}$ if they are correlated with a target. This disqualifying correlation results from the proximity of close targets. 

\begin{figure*}[!t]
\centering
\includegraphics[width=6.75in]{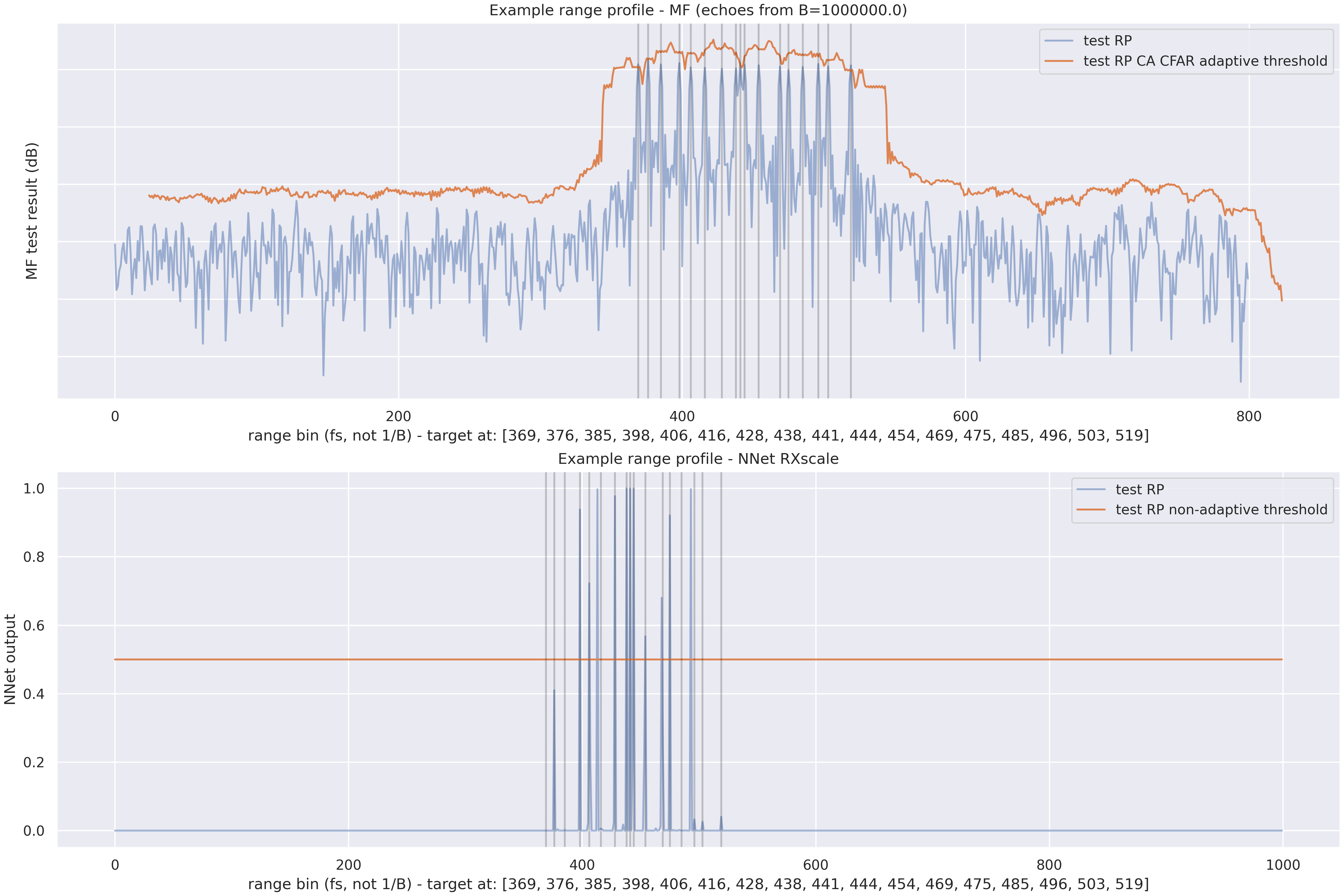}
\caption{Example range profile processing for a test range profile. From top to bottom: the first figure is the matched filtered range profile with the corresponding CA-CFAR threshold; the second figure is the neural network processed range profile with the corresponding arbitrary threshold set to $0.5$.}
\label{example_test_rp}
\end{figure*}

Example matched filter and neural network outputs are shown on Fig.~\ref{example_test_rp}. The first figure (top), corresponding to the MF, appears to be incomplete because the absence of padding prevents the computation of a valid scalar product when the distance to the end of the range profile is lower than the pulse length. On the second figure (bottom), the arbitrary threshold contributes to making the detection performances robust with respect to targets density : neighboring targets do not raise the local threshold. Since one in two range bins are used as reference or guard cells to define the local detection threshold, two neighboring range bins have their respective thresholds defined by disjoint sets of bins. Also, since this is a range profile from the test set the targets positioning distribution along the range axis is irregular, as explained in Section \ref{datasets-generation}. 

The neural network was trained during 100 epochs using the Adam optimizer, a scheduler halving the learning rate every 20 epochs, a mean squared error (MSE) loss and batches of $512$ range profiles. The MSE loss penalizes misalignments between binary labels reflecting the presence of a target within individual range bins and the neural network final sigmoid output. This loss is parameterized by class weights :  it multiplies by 10 the loss contribution of non-empty range bins, the latter being less common. Experiments were implemented with PyTorch and NumPy.


Test set detection metrics are reported in Table \ref{experiments-metrics}. The two first columns report neural network detection metrics, whereas the last column reports the performances achieved by detecting  echoes using the matched filter followed by the CA-CFAR threshold. The neural network performances reported are the average of three runs launched with distinct random generator seeds. The first column metrics correspond to the neural network trained on the "baseline" training set and the second column metrics correspond to the neural network trained on the "enriched" set, the latter including two pulse bandwidths in addition to empty and contrastive range profiles. The first line reports the overall performances computed over the entire test set, and the following lines apply a filter to the test range profiles to observe the performances computed over a subset of the test data.

The detection metrics stemming from the neural network respect the $P_{fa}=10^{-3}$ target set for the CA-CFAR threshold computation, and even reach lower false alarm levels. This makes the detection probabilities $P_d$ shown in Table \ref{experiments-metrics} comparable. This in turn allows us to conclude that the neural network approach is substantially superior to the classical matched filter followed by CA-CFAR approach on our test set. Indeed, when the training set used to optimize the neural network is the "baseline" one, the neural network remains inferior to the classical method. However, once reduced bandwidth samples, empty and contrastive range profiles are added the neural network clearly outperforms the MF+CA-CFAR approach.  One should nonetheless notice that the latter remains superior in $P_d$ alone for low energy echoes, i.e. when the reflection coefficient is set to $0.1$, while both methods remain relatively weak for the reduced bandwidth samples. The CA-CFAR adaptive threshold proves itself to be adapted to match locally reduced signal energy. The neural network can be said to not generalize very well to low amplitude, frequency-disturbed echoes. This shortcoming could probably be corrected using further extended datasets. The weakness of the matched filter with respect to the altered bandwidth samples is not surprising as the matching of frequency patterns is imperfect, shifting the peak among range bins. While not reported here, the training data metrics demonstrated equivalent $P_{fa}$ for the MF+CA-CFAR method, as expected. Overall, the seemingly low detection performances stem from the pathological setting this work intentionally selected to study a hard detection problem.

\begin{table}[htbp]
\caption{Datasets generation parameters.}
\label{data_gen_params}
\begin{center}
\begin{tabular}{|l|c|c|c|c|}
\hline
\bfseries parameter & \bfseries train. set & \bfseries train. set & \bfseries valid. set & \bfseries test. set \\
& \bfseries (baseline) & \bfseries (enriched) & & \\
\hline
LFM band(MHz) 			& 1.0 ($B_{train}$) & 0.96,\ 1.0 	& 0.97 		& 0.98,\ 1.0 \\
\hline
reflection coeff 		& 0.5, 1.0 			& 0.5, 1.0 		& 0.7 		& 0.1, 0.8 \\
\hline
gaussian noise std 		& 0.04, 0.06 		& 0.04, 0.06 	& 0.08 		& 0.1, 0.2 \\
\hline
\# range profiles 		& 206260 			& 919080 		& 206320 	& 928746 \\
\hline
\# targets in RP 		& 1-119 			& 1-119 		& 1-117 	& 1-119 \\
\hline
\# min/max stride 		& 5-50 				& 5-50 			& 5-50 		& 5-50 \\
\hline
\# empty RPs 			& 0 				& 408480 		& 103160 	& 412776 \\
\hline
\# contrastive RPs 		& 0 				& 102120 		& 51580 	& 103194 \\
\hline
\end{tabular}
\end{center}
\end{table}

\begin{table}[htbp]
\caption{Detection metrics computed over our test set, rounded. Each table cell reports a $P_d/P_{fa}$ couple.}
\label{experiments-metrics}
\begin{center}
\begin{tabular}{|l|l|l|l|}
\hline
\bfseries Subset filter & \bfseries NNet trained on & \bfseries NNet trained on &  \bfseries matched filter \\
 & \bfseries baseline data & \bfseries enriched data & \bfseries + CA-CFAR \\
\hline
All RPs 				& 0.1259 / 0.00002 	& \textbf{0.1760 / 0.00001} 	& 0.1465 / 0.0011 \\
\hline
Refl coeff = $0.1$ 	& 0.0279 / 1.3e-7 		& 0.0745 / 0.000005 	& \textbf{0.1220 / 0.0013} \\
\hline
Refl coeff = $0.8$ 	& 0.2239 / 0.00010 		& \textbf{0.2775 / 0.00007} 	& 0.1712 / 0.0014 \\
\hline
Noise std = $0.1$ 		& 0.1308 / 0.00002 	& \textbf{0.1848 / 0.000010} 	& 0.1567 / 0.0011 \\
\hline
Noise std = $0.2$ 		& 0.1210 / 0.00002 	& \textbf{0.1672 / 0.00002} 	& 0.1365 / 0.0011 \\
\hline
LFM 1 MHz				& 0.2377 / 2.3e-7 		& \textbf{0.2852 / 0.000009} 	& 0.2572 / 0.0010 \\
\hline
LFM 0.98 MHz 			& 0.0141 / 0.00010 		& \textbf{0.0668 / 0.00006} 	& 0.0360 / 0.0016 \\
\hline
\end{tabular}
\end{center}
\end{table}

The validation data revealed few training epochs are needed to reach the point where validation metrics go the wrong way, suggesting overfitting. This is not surprising as the training set diversity is limited, making the training data only evocative of the expected swarm echoes. This is deliberate as this work aims at evaluating the potential of neural networks for targets swarms detection with unpredictable waveform distortions and various clutter levels. No early stopping or dropout strategy was implemented, effectively keeping our machine learning approach very simple. 

\section{Conclusion}
\label{conclusion}

This paper put forward a partially complex-valued neural network-based range profile processing. The latter demonstrated its potential to rival the matched filter performances while being differently impacted by various pulse echo distortions. The approach proposed by this work is a natural evolution of iterative and gradient-based adaptive pulse compression and takes advantage of modern neural networks. The neural network architecture proposed is lightweight and could therefore be well suited for embedded systems, for instance radars on unmanned platforms. Future work will compare range profile-scale with waveform-scale neural network processing, investigate robustness with respect to more diverse and stronger waveform distortions, and explore training set augmentations. The relevance and accuracy of targets swarms simulated representations will be improved, as this work spread targets rather widely along the observed range. Giving the neural network replacing pulse compression access to the undistorted transmitted pulse will also be evaluated in order to reflect the definition of matched filter weights.

\bibliographystyle{IEEEtran}
\bibliography{IEEEabrv,submission_eusipco2026_mbauw_v2}

\begin{thebibliography}{10}
\providecommand{\url}[1]{#1}
\csname url@samestyle\endcsname
\providecommand{\newblock}{\relax}
\providecommand{\bibinfo}[2]{#2}
\providecommand{\BIBentrySTDinterwordspacing}{\spaceskip=0pt\relax}
\providecommand{\BIBentryALTinterwordstretchfactor}{4}
\providecommand{\BIBentryALTinterwordspacing}{\spaceskip=\fontdimen2\font plus
\BIBentryALTinterwordstretchfactor\fontdimen3\font minus
  \fontdimen4\font\relax}
\providecommand{\BIBforeignlanguage}[2]{{%
\expandafter\ifx\csname l@#1\endcsname\relax
\typeout{** WARNING: IEEEtran.bst: No hyphenation pattern has been}%
\typeout{** loaded for the language `#1'. Using the pattern for}%
\typeout{** the default language instead.}%
\else
\language=\csname l@#1\endcsname
\fi
#2}}
\providecommand{\BIBdecl}{\relax}
\BIBdecl

\bibitem{rohling1983radar}
H.~Rohling, ``Radar cfar thresholding in clutter and multiple target
  situations,'' \emph{IEEE transactions on aerospace and electronic systems},
  no.~4, pp. 608--621, 1983.

\bibitem{gandhi1988analysis}
P.~P. Gandhi and S.~A. Kassam, ``Analysis of cfar processors in nonhomogeneous
  background,'' \emph{IEEE Transactions on Aerospace and Electronic systems},
  vol.~24, no.~4, pp. 427--445, 1988.

\bibitem{rihaczek1965radar}
A.~W. Rihaczek, ``Radar signal design for target resolution,''
  \emph{Proceedings of the IEEE}, vol.~53, no.~2, pp. 116--128, 1965.

\bibitem{blunt2006adaptive}
S.~D. Blunt and K.~Gerlach, ``Adaptive pulse compression via mmse estimation,''
  \emph{IEEE Transactions on Aerospace and Electronic Systems}, vol.~42, no.~2,
  pp. 572--584, 2006.

\bibitem{rabaste2015mismatched}
O.~Rabaste and L.~Savy, ``Mismatched filter optimization for radar applications
  using quadratically constrained quadratic programs,'' \emph{IEEE Transactions
  on Aerospace and Electronic Systems}, vol.~51, no.~4, pp. 3107--3122, 2015.

\bibitem{mccormick2016gradient}
P.~M. McCormick, S.~D. Blunt, and T.~Higgins, ``A gradient descent
  implementation of adaptive pulse compression,'' in \emph{2016 IEEE Radar
  Conference (RadarConf)}.\hskip 1em plus 0.5em minus 0.4em\relax IEEE, 2016,
  pp. 1--5.

\bibitem{xu2017evaluations}
J.~Xu, B.~Bai, C.~Dong, Y.~Dong, Y.~Zhu, and G.~Zhao, ``Evaluations of plasma
  stealth effectiveness based on the probability of radar detection,''
  \emph{IEEE Transactions on Plasma Science}, vol.~45, no.~6, pp. 938--944,
  2017.

\bibitem{akhtar2018neural}
J.~Akhtar and K.~E. Olsen, ``A neural network target detector with partial
  ca-cfar supervised training,'' in \emph{2018 International Conference on
  Radar (RADAR)}.\hskip 1em plus 0.5em minus 0.4em\relax IEEE, 2018, pp. 1--6.

\bibitem{diskin2024cfarnet}
T.~Diskin, Y.~Beer, U.~Okun, and A.~Wiesel, ``Cfarnet: Deep learning for target
  detection with constant false alarm rate,'' \emph{Signal Processing}, vol.
  223, p. 109543, 2024.

\bibitem{smith2024frequency}
J.~W. Smith and M.~Torlak, ``Frequency estimation using complex-valued shifted
  window transformer,'' \emph{IEEE Geoscience and Remote Sensing Letters},
  2024.

\bibitem{lei2024understanding}
J.~Lei, Y.~Li, L.-Y. Yung, Y.~Leng, Q.~Lin, and Y.-C. Wu, ``Understanding
  complex-valued transformer for modulation recognition,'' \emph{IEEE Wireless
  Communications Letters}, 2024.

\bibitem{brooks2019hermitian}
D.~Brooks, O.~Schwander, F.~Barbaresco, J.-Y. Schneider, and M.~Cord, ``A
  hermitian positive definite neural network for micro-doppler complex
  covariance processing,'' in \emph{2019 International Radar Conference
  (RADAR)}.\hskip 1em plus 0.5em minus 0.4em\relax IEEE, 2019, pp. 1--6.

\bibitem{pervdoch2024improved}
J.~Per{\v{d}}och, S.~Ga{\v{z}}ovov{\'a}, and M.~Pacek, ``An improved radar
  clutter suppression by simple neural network,'' \emph{IET Radar, Sonar \&
  Navigation}, vol.~18, no.~2, pp. 308--326, 2024.

\bibitem{gandhi1997neural}
P.~P. Gandhi and V.~Ramamurti, ``Neural networks for signal detection in
  non-gaussian noise,'' \emph{IEEE transactions on Signal Processing}, vol.~45,
  no.~11, pp. 2846--2851, 1997.

\bibitem{akhtar2023high}
J.~Akhtar, ``High-resolution neural network processing of lfm radar pulses,''
  in \emph{ICASSP 2023-2023 IEEE International Conference on Acoustics, Speech
  and Signal Processing (ICASSP)}.\hskip 1em plus 0.5em minus 0.4em\relax IEEE,
  2023, pp. 1--5.

\bibitem{arjovsky2016unitary}
M.~Arjovsky, A.~Shah, and Y.~Bengio, ``Unitary evolution recurrent neural
  networks,'' in \emph{International conference on machine learning}.\hskip 1em
  plus 0.5em minus 0.4em\relax PMLR, 2016, pp. 1120--1128.

\bibitem{trabelsi2017deep}
C.~Trabelsi, O.~Bilaniuk, Y.~Zhang, D.~Serdyuk, S.~Subramanian, J.~F. Santos,
  S.~Mehri, N.~Rostamzadeh, Y.~Bengio, and C.~J. Pal, ``Deep complex
  networks,'' \emph{arXiv preprint arXiv:1705.09792}, 2017.

\end{thebibliography}

\end{document}